\def\year{2018}\relax
\documentclass[letterpaper]{article} 
\usepackage{aaai18}  
\usepackage{times}  
\usepackage{helvet}  
\usepackage{courier}  
\usepackage{url}  
\usepackage{graphicx}  
\usepackage{multirow}
\usepackage{amsthm}
\usepackage{amssymb}
\usepackage{amsmath}
\usepackage{color}
\usepackage{MnSymbol}
\usepackage{makecell}
\usepackage{arydshln}
\usepackage{microtype}
\usepackage[shortlabels]{enumitem}
\usepackage[dvipsnames]{xcolor}

\usepackage{booktabs}

\newcommand{\ourl}{Memory Fusion Network}
\newcommand{\ours}{MFN}
\newcommand{\sos}{System of LSTMs}

\newcommand{\um}{Multi-view Gated Memory}
\newcommand{\atn}{Delta-memory Attention Network}
\newcommand{\atns}{DMAN}

\makeatletter
\newcommand\footnoteref[1]{\protected@xdef\@thefnmark{\ref{#1}}\@footnotemark}
\makeatother

\begin{document}

%

\title{\ourl \ for Multi-view Sequential Learning}
\author{
Amir Zadeh\\
Carnegie Mellon University, USA \\
{\tt abagherz@cs.cmu.edu} \\ \And
Paul Pu Liang\\
Carnegie Mellon University, USA \\
{\tt pliang@cs.cmu.edu} \\ \And
Navonil Mazumder \\
Instituto Polit\'{e}cnico Nacional, Mexico\\
{\tt navonil@sentic.net} \\ \AND
Soujanya Poria \\
NTU, Singapore \\
{\tt sporia@ntu.edu.sg} \\ \And
Erik Cambria \\
NTU, Singapore \\
{\tt cambria@ntu.edu.sg} \\ \And
Louis-Philippe Morency \\
Carnegie Mellon University, USA\\
{\tt morency@cs.cmu.edu}\\
}
\maketitle

\begin{abstract}
Multi-view sequential learning is a fundamental problem in machine learning dealing with multi-view sequences. In a multi-view sequence, there exists two forms of interactions between different views: view-specific interactions and cross-view interactions. In this paper, we present a new neural architecture for multi-view sequential learning called the \ourl \ (\ours) that explicitly accounts for both interactions in a neural architecture and continuously models them through time. The first component of the \ours \ is called the \sos, where view-specific interactions are learned in isolation through assigning an LSTM function to each view. The cross-view interactions are then identified using a special attention mechanism called the \atn \ (\atns) and summarized through time with a \um. Through extensive experimentation, \ours \ is compared to various proposed approaches for multi-view sequential learning on multiple publicly available benchmark datasets. \ours \ outperforms all the existing multi-view approaches. Furthermore, \ours \ outperforms all current state-of-the-art models, setting new state-of-the-art results for these multi-view datasets.
\end{abstract}

\section{Introduction}
In many natural scenarios, data is collected from diverse perspectives and exhibits heterogeneous properties: each of these domains present a different view of the same data, where each view can have its own individual representation space and dynamics. Such forms of data are known as multi-view data. In a multi-view setting, each view of the data may contain some knowledge that other views do not have access to. Therefore, multiple views must be employed together in order to describe the data comprehensively and accurately. Multi-view learning has been an active area of machine learning research \cite{xu2013survey}. By exploring the consistency and complementary properties of different views, multi-view learning can be more effective, more promising, and has better generalization ability than single-view learning.

Multi-view sequential learning extends the definition of multi-view learning to manage with different views all in the form of sequential data, i.e. data that comes in the form of sequences. For example, a video clip of an orator can be partitioned into three sequential views -- text representing the spoken words, video of the speaker, and vocal prosodic cues from the audio. In multi-view sequential learning, two primary forms of interactions exist. The first form is called view-specific interactions; interactions that involve only one view. For example, learning the sentiment of a speaker based only on the sequence of spoken works. More importantly, the second form of interactions are defined across different views. These are known as cross-view interactions. Cross-view interactions span across both the different views and time -- for example a listener's backchannel response or the delayed rumble of distant lightning in the video and audio views. Modeling both the view-specific and cross-view interactions lies at the core of multi-view sequential learning.

\begin{figure*}
\begin{center}
\includegraphics[width=0.7\textwidth]{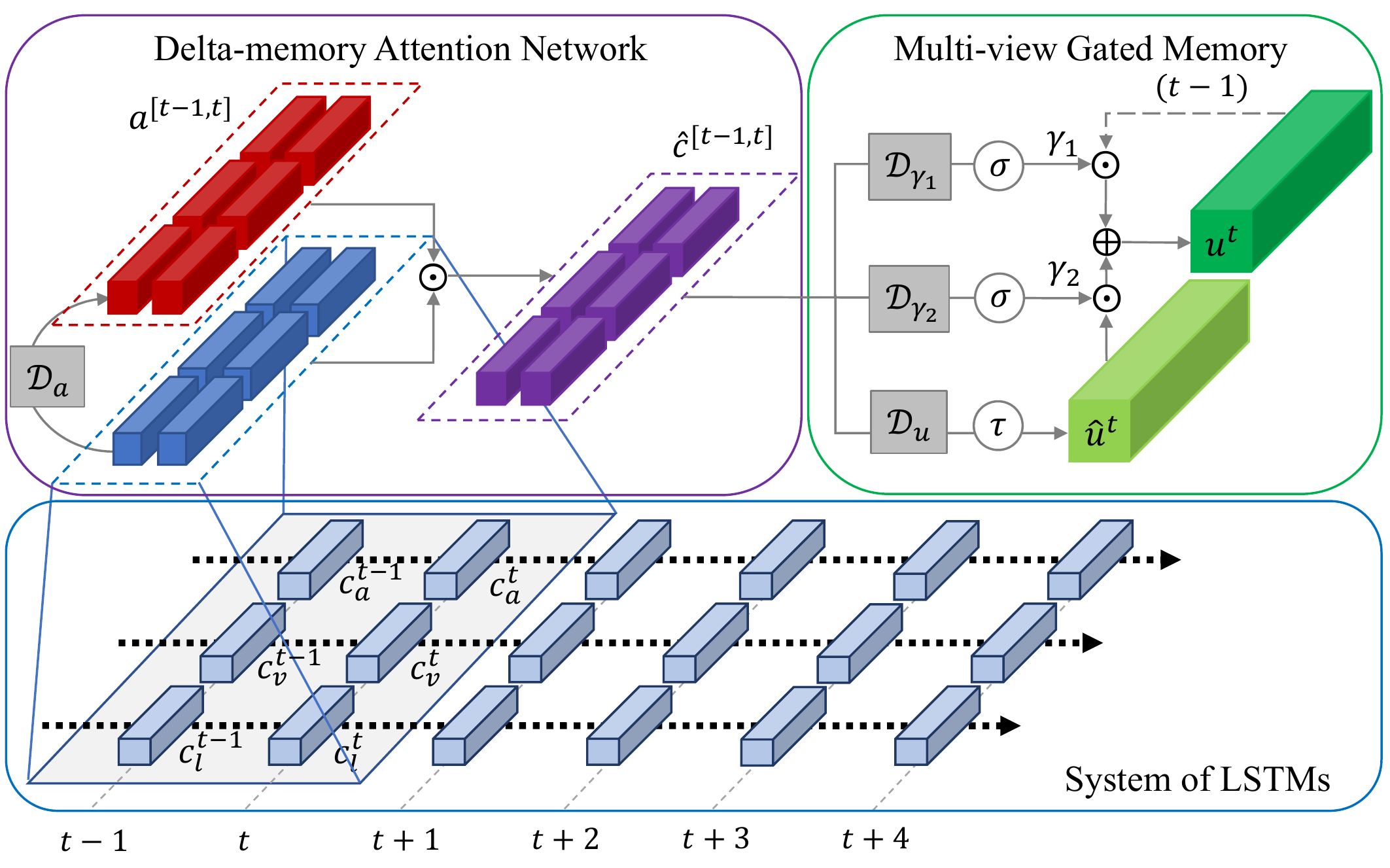}
\end{center}
\caption{\label{fig:cman} Overview figure of \ourl \ (\ours) pipeline. $\sigma$ denotes the $sigmoid$ activation function, $\tau$ the $tanh$ activation function, $\odot$ the Hadamard product and $\oplus$ element wise addition. Each LSTM encodes information from one view such as language ($l$), video ($v$) or audio ($a$).}
\end{figure*}

This paper introduces a novel neural model for multi-view sequential learning called the \ourl \ (\ours). At a first layer, the \ours \ encodes each view independently using a component called the System of Long Short Term Memories (LSTMs). In this \sos, each view is assigned one LSTM function to model the dynamics in that particular view. The second component of \ours \ is called the \atn \ (\atns) which finds cross-view interactions across memories of the \sos. Specifically, the \atns \ identifies the cross-view interactions by associating a relevance score to the memory dimensions of each LSTM. The third component of the \ours \ stores the cross-view information over time in the \um. This memory updates its contents based on the outputs of the \atns \ and its previously stored contents, acting as a dynamic memory module for learning crucial cross-view interactions throughout the sequential data. Prediction is performed by integrating both view-specific and cross-view and information.

We perform extensive experimentation to benchmark the performance of \ours \ on 6 publicly available multi-view sequential datasets. Throughout, we compare to the state-of-the-art approaches in multi-view sequential learning. In all the benchmarks, \ours \ is able to outperform the baselines, setting new state-of-the-art results across all the datasets.

\section{Related Work}
\label{Related Work}
Researchers dealing with multi-view sequential data have largely focused on three major types of models.

The first category of models have relied on concatenation of all multiple views into a single view to simplify the learning setting. These approaches then use this concatenated view as input to a learning model. Hidden Markov Models (HMMs) \cite{baum1966statistical,morency2011towards}, Support Vector Machines (SVMs) \cite{cortes1995support}, Hidden Conditional Random Fields (HCRFs) \cite{Quattoni:2007:HCR:1313053.1313265} and their variants \cite{morency2007latent} have been successfully used for structured prediction. More recently, with the advent of deep learning, Recurrent Neural Networks, specially Long-short Term Memory (LSTM) networks \cite{hochreiter1997long}, have been extensively used for sequence modeling. Some degree of success for modeling multi-view problems is achieved using this concatenation. However, this concatenation causes over-fitting in the case of a small size training sample and is not intuitively meaningful because each view has a specific statistical property \cite{xu2013survey} which is ignored in these simplified approaches.

The second category of models introduce multi-view variants to the structured learning approaches of the first category. Multi-view variations of these models have been proposed including Multi-view HCRFs where the potentials of the HCRF are changed to facilitate multiple views \cite{song2012multi,song2013action}. Recently, multi-view LSTM models have been proposed for multimodal setups where the LSTM memory is partitioned into different components for different views \cite{rajagopalan2016extending}.

The third category of models rely on collapsing the time dimension from sequences by learning a temporal representation for each of the different views. Such methods have used average feature values over time \cite{poria2015deep}. Essentially these models apply conventional multi-view learning approaches, such as Multiple Kernel Learning \cite{poria2015deep}, subspace learning or co-training \cite{xu2013survey} to the multi-view representations. Other approaches have trained different models for each view and combined the models using decision voting \cite{Nojavanasghari:2016:DMF:2993148.2993176}, tensor products \cite{tensoremnlp17} or deep neural networks \cite{contextmultimodalacl2017}. While these approaches are able to learn the relations between the views to some extent, the lack of the temporal dimension limits these learned representations, eventually affect their performance. Such is the case for long sequences where the learned representations do not sufficiently reflect all the temporal information in each view.

The proposed model in this paper is different from the first category models since it assigns one LSTM to each view instead of concatenating the information from different views. \ours \ is also different from the second category models since it considers each view in isolation to learn view-specific interactions. It then uses an explicitly designed attention mechanism and memory to find and store cross-view interactions over time. \ours \ is different from the third category models since view-specific and cross-view interactions are modeled over time. 

\section{\ourl \ (\ours)}
The \ourl \ (\ours) is a recurrent model for multi-view sequential learning that consists of three main components: 1) \textbf{\sos}\ consists of multiple Long-short Term Memory (LSTM) networks, one for each of the views. Each LSTM encodes the view-specific dynamics and interactions. 2) \textbf{\atn}\ is a special attention mechanism designed to discover both cross-view and temporal interactions across different dimensions of memories in the \sos. 3) \textbf{\um} \ is a unifying memory that stores the cross-view interactions over time. Figure \ref{fig:cman} shows the overview of \ours \ pipeline and its components.

The input to \ours \ is a multi-view sequence with the set of $N$ views each of and length $T$. For example sequences can consist of language, video, and audio for $N=\{l,v,a\}$. The input data of the $n$th view is denoted as: $\mathbf{x}_n=[x^{t}_n : t \leq T, x^{t}_n \in \mathbb{R}^{d_{x_n}}]$ where $d_{x_n}$ is the input dimensionality of $n$th view input $\mathbf{x}_n$.

\subsection{\sos}
For each view sequence, a Long-Short Term Memory (LSTM), encodes the view-specific interactions over time. At each input timestamp $t$, information from each view is input to the assigned LSTM. For the $n$th view, the memory of assigned LSTM is denoted as $\mathbf{c}_n=\{c_n^{t} : t \le T, c_n^{t} \in \mathbb{R}^{d_{c_n}}\}$ and the output of each LSTM is defined as $\mathbf{h}_n=\{h_n^{t} : t \le T, h_n^{t} \in \mathbb{R}^{d_{c_n}}\}$ with $d_{c_n}$ denoting the dimensionality of $n$th LSTM memory $\mathbf{c}_n$. Note that the \sos \ allows different sequences to have different input, memory and output shapes. The following update rules are defined for the $n$th LSTM \cite{hochreiter1997long}:

\begin{align}
i_n^{t} &= \sigma(W_n^i\ x^{t}_n + U_n^i\ h^{t-1}_n + b_n^i) \\
f_n^{t} &= \sigma(W_n^f\ x^{t}_n + U_n^f\ h^{t-1}_n + b_n^f) \\
o_n^{t} &= \sigma(W_n^o\ x^{t}_n + U_n^o\ h^{t-1}_n + b_n^o) \\
m_n^{t} &= W_n^m \ x^{t}_n + U_n^m \ h^{t-1}_n + b_n^m \\
c_n^{t}&=f_n^{t} \odot c_n^{t-1} + i_n^t \odot m_n^{t} \\
h_n^{t}&=o_n^{t} \odot tanh (c_n^{t})
\end{align}
In the above equations, the trainable parameters are the two affine transformations $W_n^* \in \mathbb{R}^{d_{x_n} \times d_{c_n}}$ and $U_n^* \in \mathbb{R}^{d_{c_n} \times d_{c_n}}$. $i_n,f_n,o_n$ are the input, forget and output gates of the $n$th LSTM respectively, $m_n$ is the proposed memory update of $n$th LSTM for time $t$, $\odot$ denotes the Hadamard product (element-wise product), $\sigma$ is the sigmoid activation function.

\subsection{\atn} 

The goal of the \atn \ (\atns) is to outline the cross-view interactions at timestep $t$ between different view memories in the \sos. To this end, we use a coefficient assignment technique on the concatenation of LSTM memories ${c}^{t}$ at time $t$. High coefficients are assigned to the dimensions jointly form a cross-view interaction and low coefficients to the other dimensions. However, coefficient assignment using only memories at time $t$ is not ideal since the same cross-view interactions can happen over multiple time instances if the LSTM memories in those dimensions remain unchanged. This is especially troublesome if the recurring dimensions are assigned high coefficients, in which case they will dominate the coefficient assignment system. To deal with this problem we add the memories ${c}^{t-1}$ of time $t-1$ so \atns \ can have the freedom of leaving unchanged dimensions in the \sos \ memories and only assign high coefficient to them if they are about to change. Ideally each cross-view interaction is only assigned high coefficients once before the state of memories in \sos \ changes. This can be done by comparing the memories at the two time-steps (hence the name Delta-memory).

The input to the \atns \ is the concatenation of memories at time $t-1$ and $t$, denoted as ${c}^{[t-1,t]}$. These memories are passed to a neural network $\mathcal{D}_a: \mathbb{R}^{2 {d_{c}}} \mapsto \mathbb{R}^{2 {d_{c}}}, d_c = \sum_n d_{c_n}$ to obtain the attention coefficients. 

\begin{equation}
{a}^{[t-1,t]} = \mathcal{D}_a ({c}^{[t-1,t]}) 
\end{equation}

${a}^{[t-1,t]}$ are softmax activated scores for each LSTM memory at time $t-1$ and $t$. Applying softmax at the output layer of $\mathcal{D}_a$ allows for regularizing high-value coefficients over the ${c}^{[t-1,t]}$. The output of the \atns \ is $\hat{c}$ defined as: 

\begin{equation}
\label{eq:chat}
\hat{c}^{[t-1,t]} = {c}^{[t-1,t]} \odot {a}^{[t-1,t]}
\end{equation}

$\hat{c}^{[t-1,t]}$ is the attended memories of the LSTMs. Applying this element-wise product amplifies the relevant dimensions of the ${c}^{[t-1,t]}$ while marginalizing the effect of remaining dimensions. \atns \ is also able to find cross-view interactions that do not happen simultaneously since it attends to the memories in the \sos. These memories can carry information about the observed inputs across different timestamps.

\subsection{\um}
\um \ $u$ is the neural component that stores a history of cross-view interactions over time. It acts as a unifying memory for the memories in \sos. The output of \atns \, $\hat{c}^{[t-1,t]}$ is directly passed to the \um \ to signal what dimensions in the \sos \ memories constitute a cross-view interaction. $\hat{c}^{[t-1,t]}$ is first used as input to a neural network $\mathcal{D}_u : \mathbb{R}^{2 \times {d_{c}}} \mapsto \mathbb{R}^{{d_{mem}}}$ to generate a cross-view update proposal $\hat{u}^{t}$ for \um. ${d_{mem}}$ is the dimensionality of the \um.
\begin{equation}
\hat{u}^{t} = \mathcal{D}_u (\hat{c}^{[t-1,t]}) 
\end{equation}

This update proposes changes to \um \ based on observations about cross-view interactions at time $t$. 

The \um \ is controlled using set of two gates. $\gamma_1$, $\gamma_2$ are called the retain and update gates respectively. At each timestep $t$, $\gamma_1$ assigns how much of the current state of the \um \ to remember and $\gamma_2$ assigns how much of the \um \ to update based on the update proposal $\hat{u}^{t}$. $\gamma_1$ and $\gamma_2$ are each controlled by a neural network. $\mathcal{D}_{\gamma_1}$, $\mathcal{D}_{\gamma_2} : \mathbb{R}^{2 \times {d_{c}}} \mapsto \mathbb{R}^{{d_{mem}}}$ control part of the gating mechanism of \um \ using $\hat{c}^{[t-1,t]}$ as input:

\begin{equation}
\gamma_1^t=\mathcal{D}_{\gamma_1}(\hat{c}^{[t-1,t]}), \ \gamma_2^t=\mathcal{D}_{\gamma_2}(\hat{c}^{[t-1,t]})
\end{equation}

At each time-step of \ours \ recursion, $u$ is updated using retain and update gates, $\gamma_1$ and $\gamma_2$, as well as the current cross-view update proposal $\hat{u}^{t}$ with the following formulation:

\begin{equation}
u^{t}=\gamma_1^{t} \odot u^{t-1} + \gamma_2^{t} \odot tanh(\hat{u}^{t})
\end{equation}

$\hat{u}^{t}$ is activated using $tanh$ squashing function to improve model stability by avoiding drastic changes to the \um. The Multi-view Gated Memory is different from LSTM memory in two ways. Firstly, the Multi-view Gated Memory has a more complex gating mechanism: both gates are controlled by neural networks while LSTM gates are controlled by a non-linear affine transformation. As a result, the Multi-view Gated Memory has superior representation capabilities as compared to the LSTM memory. Secondly, the value of the Multi-view Gated Memory does not go through a sigmoid activation in each iteration. We found that this helps in faster convergence.  

\subsection{Output of \ours}
The outputs of the \ours \ are the final state of the \um \ $u^{T}$ and the outputs of each of the $n$ LSTMs: 
$$ \mathbf{h}^T = \bigoplus_{n \in N} h_n^{T} $$ 
representing individual sequence information. $\bigoplus$ denotes vector concatenation.

\section{Experimental Setup}

In this section we design extensive experiments to evaluate the performance of \ours. We choose three multi-view domains: multimodal sentiment analysis, emotion recognition and speaker traits analysis. All benchmarks involve three views with completely different natures: language (text), vision (video), and acoustic (audio). The multi-view input signal is the video of a person speaking about a certain topic. Since humans communicate their intentions in a structured manner, there are synchronizations between intentions in text, gestures and tone of speech. These synchronizations constitute the relations between the three views. 

\subsection{Datasets}
In all the videos in the datasets described below, only one speaker is present in front of the camera.

\textbf{Sentiment Analysis}
The first domain in our experiments is multimodal sentiment analysis, where the goal is to identify a speaker's sentiment based on online video content. Multimodal sentiment analysis extends the conventional text-based definition of sentiment analysis to a multimodal setup where different views contribute to modeling the sentiment of the speaker. We use four different datasets for English and Spanish sentiment analysis in our experiments. The \textit{CMU-MOSI} dataset \cite{zadeh2016multimodal} is a collection of 93 opinion videos from online sharing websites. Each video consists of multiple opinion segments and each segment is annotated with sentiment in the range [-3,3]. The \textit{MOUD} dataset \cite{perez-rosas_utterance-level_2013} consists of product review videos in Spanish. Each video consists of multiple segments labeled to display positive, negative or neutral sentiment. To maintain consistency with previous works \cite{contextmultimodalacl2017,perez-rosas_utterance-level_2013} we remove segments with the neutral label. The \textit{YouTube} dataset \cite{morency2011towards} introduced tri-modal sentiment analysis to the research community. Multi-dimensional data from the audio, visual and textual modalities are collected in the form of 47 videos from the social media web site YouTube. The collected videos span a wide range of product reviews and opinion videos. These are annotated at the segment level for sentiment.  The \textit{ICT-MMMO} dataset \cite{wollmer2013youtube} consists of online social review videos that encompass a strong diversity in how people express opinions, annotated at the video level for sentiment. 

\textbf{Emotion Recognition} The second domain in our experiments is multimodal emotion recognition, where the goal is to identify a speakers emotions based on the speakers verbal and nonverbal behaviors. These emotions are categorized as basic emotions \cite{ekman1992argument} and continuous emotions \cite{gunes2010automatic}. We perform experiments on \textit{IEMOCAP} dataset \cite{Busso2008IEMOCAP:Interactiveemotionaldyadic}. IEMOCAP consists of 151 sessions of recorded dialogues, of which there are 2 speakers per session for a total of 302 videos across the dataset. Each segment is annotated for the presence of emotions (angry, excited, fear, sad, surprised, frustrated, happy, disappointed and neutral) as well as valence, arousal and dominance.

\textbf{Speaker Traits Analysis} 
The third domain in our experiments is speaker trait recognition based on communicative behavior of the speaker. The goal is to identify 16 different speaker traits. The \textit{POM} dataset \cite{Park:2014:CAP:2663204.2663260} contains 1,000 movie review videos. Each video is annotated for various personality and speaker traits, specifically: confident (con), passionate (pas), voice pleasant (voi), dominant (dom), credible (cre), vivid (viv), expertise (exp), entertaining (ent), reserved (res), trusting (tru), relaxed (rel), outgoing (out), thorough (tho), nervous (ner), persuasive (per) and humorous (hum). The short form of these speaker traits is indicated inside parentheses and used for the rest of this paper. 

\subsection{Sequence Features}
The chosen system of sequences are the three modalities: language, visual and acoustic. To get the exact utterance time-stamp of each word we perform forced alignment using P2FA \cite{P2FA} which allows us to align the three modalities together. Since words are considered the basic units of language we use the interval duration of each word utterance as a time-step. We calculate the expected video and audio features by taking the expectation of their view feature values over the word utterance time interval \cite{tensoremnlp17}. For each of the three modalities, we process the information from videos as follows.

\textbf{Language View} For the language view, Glove word embeddings \cite{pennington2014glove} were used to embed a sequence of individual words from video segment transcripts into a sequence of word vectors that represent spoken text. The Glove embeddings used are 300 dimensional word embeddings trained on 840 billion tokens from the common crawl dataset, resulting in a sequence of dimension $T \times d_{x_{text}}=T \times 300$ after alignment. The timing of word utterances is extracted using P2FA forced aligner. This extraction enables alignment between text, audio and video.

\textbf{Visual View} For the visual view, the library Facet \cite{emotient} is used to extract a set of visual features including facial action units, facial landmarks, head pose, gaze tracking and HOG features \cite{zhu2006fast}. These visual features are extracted from the full video segment at 30Hz to form a sequence of facial gesture measures throughout time, resulting in a sequence of dimension $T \times d_{x_{video}}=T \times 35$.

\textbf{Acoustic View} For the audio view, the software COVAREP \cite{degottex2014covarep} is used to extract acoustic features including 12 Mel-frequency cepstral coefficients, pitch tracking and voiced/unvoiced segmenting features \cite{drugman2011joint}, glottal source parameters 
\cite{childers1991vocal,drugman2012detection,alku1992glottal,alku1997parabolic,alku2002normalized}, peak slope parameters and maxima dispersion quotients \cite{kane2013wavelet}. These visual features are extracted from the full audio clip of each segment at 100Hz to form a sequence that represent variations in tone of voice over an audio segment, resulting in a sequence of dimension $T \times d_{x_{audio}}=T \times 74$ after alignment.

\subsection{Experimental Details}
The time steps in the sequences are chosen based on word utterances. The expected (average) visual and acoustic sequences features are calculated for each word utterance to ensure time alignment between all LSTMs. In all the aforementioned datasets, it is important that the same speaker does not appear in both train and test sets in order to evaluate the generalization of our approach. The training, validation and testing splits are performed so that the splits are speaker independent. The full set of videos (and segments for datasets where the annotations are at the resolution of segments) in each split is detailed in Table \ref{table:splits}. All baselines were re-trained using these video-level train-test splits of each dataset and with the same set of extracted sequence features. Training is performed on the labeled segments for datasets annotated at the segment level and on the labeled videos otherwise. All the code and data required to recreate the reported results are available at \url{https://github.com/A2Zadeh/MFN}. 

\begin{table}[!tbp]
\fontsize{7}{10}\selectfont
\setlength\tabcolsep{2.9pt}
\begin{tabular}{l : c : c : c : c : c : c}
\Xhline{3\arrayrulewidth}
Dataset       & \multicolumn{1}{c:}{CMU-MOSI} & \multicolumn{1}{c:}{ICT-MMMO} & \multicolumn{1}{c:}{YouTube} & \multicolumn{1}{c:}{MOUD} & \multicolumn{1}{c:}{IEMOCAP} & \multicolumn{1}{c}{POM} \\
Level       & \multicolumn{1}{c:}{Segment} & \multicolumn{1}{c:}{Video} & \multicolumn{1}{c:}{Segment} & \multicolumn{1}{c:}{Segment} & \multicolumn{1}{c:}{Segment} & \multicolumn{1}{c}{Video} \\ \Xhline{0.5\arrayrulewidth}
\# Train	& 52$\rightarrow$1284  & 220 & 30$\rightarrow$169 & 49$\rightarrow$243 & 5$\rightarrow$6373 & 600\\  
\# Valid	& 10$\rightarrow$229  & 40 & 5$\rightarrow$41 & 10$\rightarrow$37 & 1$\rightarrow$1775 & 100\\  
\# Test	& 31$\rightarrow$686  & 80 & 11$\rightarrow$59 & 20$\rightarrow$106 & 1$\rightarrow$1807 & 203\\  \Xhline{3\arrayrulewidth}
\end{tabular}
\caption{Data splits to ensure speaker independent learning. Arrows indicate the number of annotated segments in each video.}
\label{table:splits}
\end{table}

\subsection{Baseline Models}
\label{subsec:datasets}

\newcolumntype{K}[1]{>{\centering\arraybackslash}p{#1}}
\definecolor{gg}{RGB}{45,190,45}

\begin{table*}[!htbp]
\fontsize{7}{10}\selectfont
\setlength\tabcolsep{3.2pt}
\begin{tabular}{l : *{5}{K{0.995cm}} : *{4}{K{0.995cm}} : *{2}{K{0.995cm}} : *{2}{K{0.995cm}}}
\Xhline{3\arrayrulewidth}
Task			& \multicolumn{5}{c:}{CMU-MOSI Sentiment} & \multicolumn{4}{c:}{ICT-MMMO Sentiment} & \multicolumn{2}{c:}{YouTube Sentiment} & \multicolumn{2}{c}{MOUD Sentiment} \\
Metric        	& BA & F1 & MA$(7)$ & MAE & $r$ & BA & F1 & MAE & $r$ & MA$(3)$ & F1 & BA & F1 \\ \Xhline{0.5\arrayrulewidth}
SOTA2		& 73.9$^\dagger$ & 74.0$^\diamond$ & 32.4$^\S$ & 1.023$^\S$ & 0.601$^\diamond$ & 81.3$^\#$ & 79.6$^\#$ & 0.968$^\flat$ & 0.499$^\flat$ & 49.2$^\bullet$ & 49.2$^\bullet$ & 72.6$^\dagger$ & 72.9$^\dagger$\\ 
SOTA1		& 74.6$^\ast$ & 74.5$^\ast$ & 33.2$^\diamond$ & 1.019$^\diamond$ & 0.622$^\S$ & 81.3$^\blacksquare$ & 79.6$^\blacksquare$ & 0.842$^\S$ & 0.588$^\S$ & 50.2$^\clubsuit$ & 50.8$^\clubsuit$ & 74.0$^\clubsuit$ & 74.7$^\clubsuit$\\  \Xhline{0.5\arrayrulewidth}
{\ours } $l$ & 73.2 & 73.0 & 32.9 & 1.012 & 0.607 & 60.0 & 55.3 & 1.144 & 0.042 & 50.9 & 49.1 & 69.8 & 69.9 \\
{\ours } $a$ & 53.1 & 47.5 & 15.0 & 1.446 & 0.186 & 80.0 & 79.3 & 1.089 & 0.462 & 39.0 & 27.0 & 60.4 & 47.1\\
{\ours } $v$ & 55.4 & 54.7 & 15.0 & 1.446 & 0.155 & 58.8 & 58.6 & 1.204 & 0.402 & 42.4 & 35.7 & 61.3 & 47.6 \\
{\ours } (no $\Delta$)  & 75.5 & 75.2 & 34.5 & 0.980 & 0.626 & 76.3 & 75.8 & 0.890 & 0.577 & 55.9 & 55.4 & 71.7 & 70.6 \\
{\ours } (no mem)  & {76.5} & {76.5} & 30.8 	&  {0.998}   	& 0.582 & 82.5 & 82.4 & 0.883 & 0.597 & 47.5	&	42.8 & 75.5 & 72.9 \\
{\ours } 	& \textbf{77.4}	 & \textbf{77.3}   & \textbf{34.1} 	&  \textbf{0.965}    	& \textbf{0.632} & \textbf{87.5}	& \textbf{87.1} & \textbf{0.739} & \textbf{0.696} &  \textbf{61.0}	&	\textbf{60.7} & \textbf{81.1} & \textbf{80.4} \\ \Xhline{0.5\arrayrulewidth}
$\Delta_{SOTA}$	& \textcolor{gg}{$\uparrow $ \textbf{2.8}} & \textcolor{gg}{$\uparrow $ \textbf{2.8}} & \textcolor{gg}{$\uparrow $ \textbf{0.9}} & \textcolor{gg}{$\downarrow $ \textbf{0.054}} & \textcolor{gg}{$\uparrow $ \textbf{0.010}} & \textcolor{gg}{$\uparrow $ \textbf{6.2}} & \textcolor{gg}{$\uparrow $ \textbf{7.5}} & \textcolor{gg}{$\downarrow $ \textbf{0.103}} & \textcolor{gg}{$\uparrow $ \textbf{0.108}} & \textcolor{gg}{$\uparrow $ \textbf{10.8}} & \textcolor{gg}{$\uparrow $ \textbf{9.9}} & \textcolor{gg}{$\uparrow $ \textbf{7.1}} & \textcolor{gg}{$\uparrow $ \textbf{5.7}} \\ \Xhline{3\arrayrulewidth}
\end{tabular}
\label{sentiment}
\end{table*}

\begin{table*}[!htbp]
\fontsize{7}{10}\selectfont
\setlength\tabcolsep{3.2pt}
\begin{tabular}{l : *{8}{K{1.763cm}}}
\Xhline{3\arrayrulewidth}
Task			& \multicolumn{2}{c}{IEMOCAP Discrete Emotions} & \multicolumn{2}{c}{IEMOCAP Valence} & \multicolumn{2}{c}{IEMOCAP Arousal} & \multicolumn{2}{c}{IEMOCAP Dominance} \\
Metric        	& MA$(9)$ & F1 & MAE & $r$ & MAE & $r$ & MAE & $r$ \\ \Xhline{0.5\arrayrulewidth}
SOTA2		& 35.9$^\dagger$ & 34.1$^\dagger$ & 0.248$^\dagger$ & 0.065$^\dagger$ & 0.521$^\ast$ & 0.617$^\S$ & 0.671$^\ast$ & 0.479$^\S$ \\ 
SOTA1		&36.0$^\ast$ & 34.5$^\ast$ & 0.244$^\S$ & 0.088$^\S$ & 0.513$^\diamond$ & 0.620$^\diamond$ & 0.668$^\diamond$ & \textbf{0.519$^\diamond$} \\  \Xhline{0.5\arrayrulewidth}
{\ours } $l$ & 25.8 & 16.1 & 0.250 & -0.022 & 1.566 & 0.105 & 1.599 & 0.162\\
{\ours } $a$ & 22.5 & 11.6 & 0.279 & 0.034 & 1.924 & 0.447 & 1.848 & 0.417\\
{\ours } $v$ & 21.5 & 10.5 & 0.248 & -0.014 & 2.073 & 0.155 & 2.059 & 0.083\\
{\ours } (no $\Delta$) & 34.8 & 33.1 & 0.243 & 0.098 & 0.500 & 0.590 & 0.629 & 0.466\\
{\ours } (no mem)  & 31.2 & 28.0 & 0.246 & 0.089 & 0.509 & 0.634 & 0.679 & 0.441 \\
{\ours } 	& \textbf{36.5} & \textbf{34.9} & \textbf{0.236} & \textbf{0.111} & \textbf{0.482} & \textbf{0.645} & \textbf{0.612} & 0.509 \\ \Xhline{0.5\arrayrulewidth}
$\Delta_{SOTA}$	& \textcolor{gg}{$\uparrow $ \textbf{0.5}} & \textcolor{gg}{$\uparrow $ \textbf{0.4}} & \textcolor{gg}{$\downarrow $ \textbf{0.008}} & \textcolor{gg}{$\uparrow $ \textbf{0.023}} & \textcolor{gg}{$\downarrow $ \textbf{0.031}} & \textcolor{gg}{$\uparrow $ \textbf{0.025}} & \textcolor{gg}{$\downarrow $ \textbf{0.056}} & \textcolor{gray}{$\downarrow $ \textbf{0.010}} \\ \Xhline{3\arrayrulewidth}
\end{tabular}
\label{emotion}
\end{table*}

\begin{table*}[!htbp]
\fontsize{7}{10}\selectfont
\setlength\tabcolsep{3.2pt}
\begin{tabular}{l : *{16}{K{0.77cm}}}
\Xhline{3\arrayrulewidth}
Dataset & \multicolumn{16}{c}{POM} \\
Task & \multicolumn{1}{c}{Con} & \multicolumn{1}{c}{Pas} & \multicolumn{1}{c}{Voi} & \multicolumn{1}{c}{Dom} & \multicolumn{1}{c}{Cre} & \multicolumn{1}{c}{Viv} & \multicolumn{1}{c}{Exp} & \multicolumn{1}{c}{Ent} & \multicolumn{1}{c}{Res} & \multicolumn{1}{c}{Tru} & \multicolumn{1}{c}{Rel} & \multicolumn{1}{c}{Out} & \multicolumn{1}{c}{Tho} & \multicolumn{1}{c}{Ner} & \multicolumn{1}{c}{Per} & \multicolumn{1}{c}{Hum}\\
Metric & MA$(7)$ & MA$(7)$ & MA$(7)$ & MA$(7)$ & MA$(7)$ &  MA$(7)$ &  MA$(7)$ &  MA$(7)$ & MA$(5)$ & MA$(5)$ & MA$(5)$ & MA$(5)$ &  MA$(5)$ &  MA$(5)$ & MA$(7)$ & MA$(5)$  \\ 
\Xhline{0.5\arrayrulewidth}
SOTA2		& 26.6$^\bullet$&27.6$^\S$&32.0$^\heartsuit$&35.0$^\heartsuit$&26.1$^\flat$&32.0$^\flat$&27.6$^\ast$&29.6$^\flat$&34.0$^\heartsuit$&53.2$^\bullet$&49.8$^\heartsuit$&39.4$^\flat$&42.4$^\S$&42.4$^\flat$&27.6$^\ast$&36.5$^\dagger$ \\
SOTA1		&  26.6$^\bullet$&31.0$^\ast$&33.0$^\flat$&35.0$^\heartsuit$&27.6$^\dagger$&36.5$^\dagger$&30.5$^\dagger$&31.5$^\heartsuit$&34.0$^\heartsuit$&53.7$^\flat$&50.7$^\diamond$&42.9$^\heartsuit$&45.8$^\dagger$&42.4$^\flat$&28.1$^\heartsuit$&40.4$^\bullet$ \\  \Xhline{0.5\arrayrulewidth}
{\ours } $l$ & 26.6 & 31.5 & 21.7 & 34.0 & 25.6 & 28.6 & 26.6 & 30.5 & 29.1 & 34.5 & 39.9 & 31.5 & 30.5 & 34.0 & 24.1 & 42.4  \\
{\ours } $a$ & 27.1&26.1&29.6&34.5&24.6&29.6&26.6&31.0&32.5&35.0&45.8&37.4&35.0&40.4&28.1&36.5 \\
{\ours } $v$ & 25.6&23.6&26.6&31.5&25.1&28.6&25.6&26.6&32.5&48.3&43.3&36.9&42.4&33.5&24.1&37.4\\
{\ours } (no $\Delta$) & 28.1&32.0&34.5&36.0&32.0&33.0&29.6&33.5&33.0&56.2&51.2&42.9&44.3&43.8&31.5&42.9 \\
{\ours } (no mem) 	&  26.1&27.1&34.5&35.5&28.1&31.0&27.1&30.0&32.0&55.2&50.7&39.4&42.9&42.4&29.1&33.5\\ 
{\ours } 	& 
\textbf{34.5}&\textbf{35.5}&\textbf{37.4}&\textbf{41.9}&\textbf{34.5}&\textbf{36.9}&\textbf{36.0}&\textbf{37.9}&\textbf{38.4}&\textbf{57.1}&\textbf{53.2}&\textbf{46.8}&\textbf{47.3}&\textbf{47.8}&\textbf{34.0}&\textbf{47.3} \\ \Xhline{0.5\arrayrulewidth}
$\Delta_{SOTA}$	& \textcolor{gg}{$\uparrow $ \textbf{7.9}} & \textcolor{gg}{$\uparrow $ \textbf{4.5}} & \textcolor{gg}{$\uparrow $ \textbf{4.4}} & \textcolor{gg}{$\uparrow $ \textbf{6.9}} & \textcolor{gg}{$\uparrow $ \textbf{6.9}} & \textcolor{gg}{$\uparrow $ \textbf{0.4}} & \textcolor{gg}{$\uparrow $ \textbf{5.5}} & \textcolor{gg}{$\uparrow $ \textbf{6.4}} & \textcolor{gg}{$\uparrow $ \textbf{4.4}} &\textcolor{gg}{$\uparrow $ \textbf{3.4}} & \textcolor{gg}{$\uparrow $ \textbf{2.5}} & \textcolor{gg}{$\uparrow $ \textbf{3.9}} & \textcolor{gg}{$\uparrow $ \textbf{1.5}} & \textcolor{gg}{$\uparrow $ \textbf{5.4}}& \textcolor{gg}{$\uparrow $ \textbf{5.9}} & \textcolor{gg}{$\uparrow $ \textbf{6.9}} \\  
\Xhline{3\arrayrulewidth}
\\
\Xhline{3\arrayrulewidth}
Metric & \multicolumn{16}{c}{MAE} \\ 
\Xhline{0.5\arrayrulewidth}
SOTA2		& 1.033$^\flat$&1.067$^\S$&0.911$^\S$&0.864$^\ast$&1.022$^\S$&0.981$^\S$&0.990$^\S$&0.967$^\flat$&0.884$^\flat$&0.556$^\S$&0.594$^\S$&0.700$^\S$&0.712$^\S$&0.705$^\dagger$&1.084$^\S$&0.768$^\flat$ \\ 
SOTA1		& 1.016$^\dagger$&1.008$^\dagger$&0.899$^\flat$&0.859$^\dagger$&0.942$^\dagger$&\textbf{0.905$^\dagger$}&0.906$^\dagger$&0.927$^\dagger$&0.877$^\diamond$&0.523$^\diamond$&0.591$^\flat$&0.698$^\flat$&0.680$^\dagger$&0.687$^\diamond$&1.025$^\dagger$&0.767$^\dagger$ \\  \Xhline{0.5\arrayrulewidth}
{\ours } $l$ & 1.065 & 1.152 & 1.033 & 0.875 & 1.074 & 1.111 & 1.135 & 0.994 & 0.915 & 0.591 & 0.612 & 0.792 & 0.753 & 0.722 & 1.134 & 0.838\\
{\ours } $a$ &1.086&1.147&0.937&0.887&1.104&1.028&1.075&1.009&0.882&0.589&0.611&0.719&0.759&0.697&1.159&0.783 \\
{\ours } $v$ & 1.083&1.153&1.009&0.931&1.085&1.073&1.135&1.028&0.929&0.664&0.682&0.771&0.770&0.773&1.138&0.793\\
{\ours } (no $\Delta$) & 1.015&1.061&0.891&0.859&0.994&0.958&1.000&0.955&0.875&0.527&0.583&0.691&0.711&0.691&1.052&0.750\\
{\ours } (no mem) 	&  1.018&1.077&0.887&0.865&1.014&0.995&1.012&0.959&0.877&0.530&0.581&0.701&0.719&0.694&1.063&0.764\\  
{\ours } 	& 
\textbf{0.952}&\textbf{0.993}&\textbf{0.882}&\textbf{0.835}&\textbf{0.903}&0.908&\textbf{0.886}&\textbf{0.913}&\textbf{0.821}&\textbf{0.521}&\textbf{0.566}&\textbf{0.679}&\textbf{0.665}&\textbf{0.654}&\textbf{0.981}&\textbf{0.727} \\  \Xhline{0.5\arrayrulewidth}
$\Delta_{SOTA}$	& \textcolor{gg}{$\downarrow $ \textbf{0.064}} & \textcolor{gg}{$\downarrow $ \textbf{0.015}} & \textcolor{gg}{$\downarrow $ \textbf{0.017}} & \textcolor{gg}{$\downarrow $ \textbf{0.024}} & \textcolor{gg}{$\downarrow $ \textbf{0.039}} & \textcolor{gray}{$\uparrow $ \textbf{0.003}} & \textcolor{gg}{$\downarrow $ \textbf{0.020}} & \textcolor{gg}{$\downarrow $ \textbf{0.014}} &\textcolor{gg}{$\downarrow $ \textbf{0.056}} & \textcolor{gg}{$\downarrow $ \textbf{0.002}} & \textcolor{gg}{$\downarrow $ \textbf{0.025}} & \textcolor{gg}{$\downarrow $ \textbf{0.019}} & \textcolor{gg}{$\downarrow $ \textbf{0.015}} & \textcolor{gg}{$\downarrow $ \textbf{0.033}} & \textcolor{gg}{$\downarrow $ \textbf{0.044}} & \textcolor{gg}{$\downarrow $ \textbf{0.040}}\\ 
\Xhline{3\arrayrulewidth}
\\
\Xhline{3\arrayrulewidth}
Metric  & \multicolumn{16}{c}{$r$} \\ \Xhline{0.5\arrayrulewidth}
SOTA2		& 
0.240$^\flat$&0.302$^\S$&0.031$^\S$&0.139$^\flat$&0.170$^\S$&0.244$^\S$&0.265$^\S$&0.240$^\S$&0.148$^\flat$&0.109$^\dagger$&0.083$^\S$&0.093$^\flat$&0.260$^\S$&0.136$^\flat$&0.217$^\S$&0.259$^\flat$ \\ 
SOTA1		& 0.359$^\dagger$&0.425$^\dagger$&0.131$^\diamond$&0.234$^\dagger$&0.358$^\dagger$&0.417$^\dagger$&0.450$^\dagger$&0.361$^\dagger$&0.295$^\diamond$&0.237$^\diamond$&0.119$^\diamond$&0.238$^\diamond$&0.363$^\dagger$&0.258$^\diamond$&0.344$^\dagger$&0.319$^\dagger$
\\ \Xhline{0.5\arrayrulewidth}
{\ours } $l$ & 0.223 & 0.281 & -0.013 & 0.118 & 0.141 & 0.189 & 0.188 & 0.227 & -0.168 & -0.064 & 0.126 & 0.095 & 0.173 & 0.024 & 0.183 & 0.216 \\
{\ours } $a$ & 0.092&0.128&-0.019&0.050&0.021&-0.007&0.035&0.130&0.152&-0.071&0.019&-0.003&-0.019&0.106&0.024&0.064\\
{\ours } $v$ & 0.146&0.091&-0.077&-0.012&0.019&-0.035&0.012&0.038&-0.004&-0.169&0.030&-0.026&0.047&0.059&0.078&0.159\\
{\ours } (no $\Delta$) & 0.307&0.373&0.140&0.209&0.272&0.334&0.333&0.305&0.194&0.218&0.160&0.152&0.277&0.182&0.288&0.334\\ 
{\ours } (no mem)	&  0.259&0.261&0.166&0.109&0.161&0.188&0.209&0.247&0.189&0.059&0.151&0.115&0.161&0.134&0.190&0.231\\ 
{\ours }  &  \textbf{0.395}&\textbf{0.428}&\textbf{0.193}&\textbf{0.313}&\textbf{0.367}&\textbf{0.431}&\textbf{0.452}&\textbf{0.395}&\textbf{0.333}&\textbf{0.296}&\textbf{0.255}&\textbf{0.259}&\textbf{0.381}&\textbf{0.318}&\textbf{0.377}&\textbf{0.386}\\ \Xhline{0.5\arrayrulewidth}
$\Delta_{SOTA}$	& \textcolor{gg}{$\uparrow $ \textbf{0.036}} & \textcolor{gg}{$\uparrow $ \textbf{0.003}} & \textcolor{gg}{$\uparrow $ \textbf{0.062}} & \textcolor{gg}{$\uparrow $ \textbf{0.079}} & \textcolor{gg}{$\uparrow $ \textbf{0.009}} & \textcolor{gg}{$\uparrow $ \textbf{0.014}} & \textcolor{gg}{$\uparrow $ \textbf{0.002}} & \textcolor{gg}{$\uparrow $ \textbf{0.034}} & \textcolor{gg}{$\uparrow $ \textbf{0.038}} &\textcolor{gg}{$\uparrow $ \textbf{0.059}} & \textcolor{gg}{$\uparrow $ \textbf{0.136}} & \textcolor{gg}{$\uparrow $ \textbf{0.021}} & \textcolor{gg}{$\uparrow $ \textbf{0.018}} & \textcolor{gg}{$\uparrow $ \textbf{0.060}}& \textcolor{gg}{$\uparrow $ \textbf{0.033}} & \textcolor{gg}{$\uparrow $ \textbf{0.067}} \\ 
\Xhline{3\arrayrulewidth}

\end{tabular}
\caption{Results for sentiment analysis on CMU-MOSI, ICT-MMMO, YouTube and MOUD, emotion recognition on IEMOCAP and personality trait recognition on POM. SOTA1 and SOTA2 refer to the previous best and second best state of the art respectively. Best results are highlighted in bold, $\Delta_{SOTA}$ shows the change in performance over SOTA1. Improvements are highlighted in green. The \ours \ significantly outperforms SOTA across all datasets and metrics except $\Delta_{SOTA}$ entries in gray.}
\label{table:overall-4}
\end{table*}

We compare the performance of the \ours \ with current state-of-the-art models for multi-view sequential learning. To perform a more extensive comparison we train all the following baselines across all the datasets. Due to space constraints, each baseline name is denoted by a symbol (in parenthesis) which is used in Table \ref{table:overall-4} to refer to specific baseline results.

\subsubsection{View Concatenation Sequential Learning Models} \ \\
\indent \textit{Song2013} ($\lhd$): This is a layered model that uses CRFs with latent variables to learn hidden spatio-temporal dynamics. For each layer an abstract feature representation is learned through non-linear gate functions. This procedure is repeated to obtain a hierarchical sequence summary (HSS) representation \cite{song2013action}. 

\textit{Morency2011} ($\times$): Hidden Markov Model is a statistical Markov model in which the system being modeled is assumed to be a Markov process with unobserved (i.e. hidden) states \cite{baum1966statistical}. We follow the implementation in \cite{morency2011towards} for tri-modal data. 

\textit{Quattoni2007} ($\wr$): Concatenated features are used as input to a Hidden Conditional Random Field (HCRF) \cite{Quattoni:2007:HCR:1313053.1313265}. HCRF learns a set of latent variables conditioned on the concatenated input at each time step.

\textit{Morency2007} ($\#$): Latent Discriminative Hidden Conditional Random Fields (LDHCRFs) are a class of models that learn hidden states in a Conditional Random Field using a latent code between observed input and hidden output  \cite{morency2007latent}. 

\textit{Hochreiter1997} ($\S$): A LSTM with concatenation of data from different views as input \cite{hochreiter1997long}. Stacked, bidirectional and stacked bidirectional LSTMs are also trained in a similar fashion for stronger baselines. 

\subsubsection{Multi-view Sequential Learning Models}
\ \\
\indent \textit{Rajagopalan2016} ($\diamond$): Multi-view (MV) LSTM \cite{rajagopalan2016extending} aims to extract information from multiple sequences by modeling sequence-specific and cross-sequence interactions over time and output. It is a strong tool for synchronizing a system of multi-dimensional data sequences.

\textit{Song2012} ($\rhd$): MV-HCRF \cite{song2012multi} is an extension of the HCRF for Multi-view data. Instead of view concatenation, view-shared and view specific sub-structures are explicitly learned to capture the interaction between views. We also implement the topological variations - linked, coupled and linked-couple that differ in the types of interactions between the modeled views. \textit{Song2012LD} ($\blacksquare$): is a variation of this model that uses LDHCRF instead of HCRF. 

\textit{Song2013MV} ($\cup$): MV-HSSHCRF is an extension of \textit{Song2013} that performs Multi-view hierarchical sequence summary representation. 

\subsubsection{Dataset Specific Baselines}
\ \\
\indent \textit{Poria2015} ($\clubsuit$): Multiple Kernel Learning \cite{bach2004multiple} classifiers have been widely applied to problems involving multi-view data. Our implementation follows a previously proposed model for multimodal sentiment analysis \cite{poria2015deep}. 

\textit{Nojavanasghari2016} ($\flat$): Deep Fusion Approach \cite{Nojavanasghari:2016:DMF:2993148.2993176} trains single neural networks for each view's input and combine the views with a joint neural network. This baseline is current state of the art in POM dataset. 

\textit{Zadeh2016} ($\heartsuit$): Support Vector Machine \cite{cortes1995support} is a widely used classifier. This baseline is closely implemented similar to a previous work in multimodal sentiment analysis \cite{zadeh2016multimodal}. 

\textit{Ho1998} ($\bullet$): We also compare to a Random Forest \cite{ho1998random} baseline as another strong non-neural classifier.

\subsubsection{Dataset Specific State-of-the-art Baselines}
\ \\
\indent \textit{Poria2017} ($\dagger$): Bidirectional Contextual LSTM \cite{contextmultimodalacl2017} performs context-dependent fusion of multi-sequence data that holds the state of the art for emotion recognition on IEMOCAP dataset and sentiment analysis on MOUD dataset. 

\textit{Zadeh2017} ($\ast$): Tensor Fusion Network \cite{tensoremnlp17} learns explicit uni-view, bi-view and tri-view concepts in multi-view data. It is the current state of the art for sentiment analysis on CMU-MOSI dataset. 

\textit{Wang2016} ($\cap$): Selective Additive Learning Convolutional Neural Network \cite{wang2016select} is a multimodal sentiment analysis model that attempts to prevent identity-dependent information from being learned so as to improve generalization based only on accurate indicators of sentiment. 

\subsubsection{\ours \ Ablation Study Baselines} 
\ \\
\indent {\ours \ $\{l,v,a\}$}: These baselines use only individual views -- $l$ for language, $v$ for visual, and $a$ for acoustic. The \atns \ and \um \ are also removed since only one view is present. This effectively reduces the \ours \ to one single LSTM which uses input from one view. 

{\ours \ (no $\Delta$)}: This variation of our model shrinks the context to only the current timestamp $t$ in the \atns. We compare to this model to show the importance of having the $\Delta$ memory temporal information -- memories at both time $t$ and $t-1$.

{\ours \ (no mem)}: This variation of our model removes the \atn \ and \um \ from the \ours. Essentially this is equivalent to three disjoint LSTMs. The output of the \ours \ in this case would only be the outputs of LSTM at the final timestamp $T$. This baseline is designed to evaluate the importance of spatio-temporal relations between views through time. 

\section{\ours \ Results and Discussion}
Table \ref{table:overall-4} summarizes the comparison between \ours \ and proposed baselines for sentiment analysis, emotion recognition and speaker traits recognition. Different evaluation tasks are performed for different datasets based on the provided labels: binary classification, multi-class classification, and regression. For binary classification we report results in binary accuracy (BA) and binary F1 score. For multiclass classification we report multiclass accuracy MA$(k)$ where $k$ denotes the number of classes, and multiclass F1 score. For regression we report Mean Absolute Error (MAE) and Pearson's correlation $r$. Higher values denote better performance for all metrics. The only exception is MAE which lower values indicate better performance. All the baselines are trained for all the benchmarks using the same input data as \ours \ and best set of hyperparameters are chosen based on a validation set according to Table \ref{table:splits}. The best performing baseline for each benchmark is referred to as state of the art 1 (SOTA1) and SOTA2 is the second best performing model. SOTA models change across different metrics since different models are suitable for different tasks. The superscript symbol on each number indicates what method it belongs to. The performance improvement of our \ours \ over the SOTA1 model is denoted as $\Delta_{SOTA}$, the raw improvement over the previous models. The results of our experiments can be summarized as follows: 
\newline 
\textbf{\ours \ Achieves State-of-The-Art Performance for Multi-view Sequential Modeling:} Our approach significantly outperforms the proposed baselines, setting new state of the art in all datasets. Furthermore, \ours \ shows a consistent trend for both classification and regression. The same is not true for other baselines as their performance varies based on the dataset and evaluation task. Additionally, the better performance of \ours \ is not at the expense of higher number of parameters or lower speed: the most competitive baseline in most datasets is \textit{Zadeh2017} which contains roughly 2e7 parameters while \ours \ contains roughly 5e5 parameters. On a Nvidia GTX 1080 Ti GPU, \textit{Zadeh2017} runs with an average frequency of 278 IPS (data point inferences per second) while our model runs at an ultra realtime frequency of 2858 IPS.
\newline
\textbf{Ablation Studies:} Our comparison with variations of our model show a consistent trend: 
\begin{equation*}
\textrm{\ours}>\textrm{\ours \ (no }\Delta\textrm{)}, \textrm{\ours \ (no mem)}>\textrm{\ours \ $\{l,v,a\}$} 
\end{equation*}
The comparison between $\textrm{\ours \ and \ours \ (no $\Delta$)}$ indicates the crucial role of the memories of time $t-1$. The comparison between \ours \ and \ours \ (no mem) shows the essential role of the \um. The final observation comes from comparing all multi-view variations of \ours \ with single view \ours \ $\{l,v,a\}$. This indicates that using multiple views results in better performance even if various crucial components are removed from \ours. 
\newline 
\textbf{Increasing The \atns \ Input Region Size:} In our set of experiments increasing the $\Delta$ to cover $[t-q,t]$ instead of $[t-1,t]$ did not significantly improve the performance of the model. We argue that this is because additional memory steps do not add any information to the \atns \ internal mechanism.

\section{Conclusion}
This paper introduced a novel approach for multi-view sequential learning called \ourl \ (\ours). The first component of \ours \ is called \sos. In \sos, each view is assigned one LSTM function to model the interactions within the view. The second component of \ours \  is called \atn \ (\atns). \atns \ outlines the relations between views through time by associating a cross-view relevance score to the memory dimensions of each LSTM. The third component of the \ours \ unifies the sequences and is called \um. This memory updates its content based on the outputs of \atns \ calculated over memories in \sos. Through extensive experimentation on multiple publicly available datasets, the performance of \ours \ is compared with various baselines. \ours \ shows state-of-the-art performance in multi-view sequential learning on all the datasets. 

\section{Acknowledgements}
This project was partially supported by Oculus research grant. We thank the reviewers for their valuable feedback.


\bibliographystyle{aaai}
{\small\bibliography{citations.bib}}

\end{document}